\documentclass[11pt,a4paper]{article}
\usepackage[hyperref]{acl2021}
\usepackage{times}
\usepackage{latexsym}
\usepackage{multirow}
\usepackage{amsmath}
\usepackage{amsfonts}
\usepackage{graphicx}
\usepackage{booktabs}
\usepackage{diagbox}
\usepackage{amssymb}
\usepackage{makecell}
\usepackage{algorithm}  
\usepackage{algorithmicx}
\usepackage{algpseudocode}
\usepackage{amssymb}
\usepackage{times}
\usepackage{enumerate}
\usepackage{latexsym}
\usepackage{soul}
\usepackage{color}
\usepackage{verbatim}
\usepackage{tablefootnote}

\usepackage{microtype}

\aclfinalcopy 
\def\aclpaperid{3348} 

\title{Instantaneous Grammatical Error Correction with \\ Shallow Aggressive Decoding}

\author{Xin Sun$^1$\thanks{\ \ This work was done during the author's internship at MSR Asia. Contact person: Tao Ge (\href{mailto:tage@microsoft.com}{tage@microsoft.com})}~\thanks{\ \ Co-first authors with equal contributions}~~~~Tao Ge$^2$\footnotemark[2]~~~~Furu Wei$^2$~~~~Houfeng Wang$^1$ \\
$^1$ MOE Key Lab of Computational Linguistics, School of EECS, Peking University; \\
$^2$ Microsoft Research Asia  \\
{\tt \{sunx5,wanghf\}@pku.edu.cn}; \\ {\tt \{tage,fuwei\}@microsoft.com}}

\date{}

\begin{document}
\maketitle
\begin{abstract}
In this paper, we propose Shallow Aggressive Decoding (SAD) to improve the online inference efficiency of the Transformer for instantaneous Grammatical Error Correction (GEC). SAD optimizes the online inference efficiency for GEC by two innovations: 1) it aggressively decodes as many tokens as possible in parallel instead of always decoding only one token in each step to improve computational parallelism; 2) it uses a shallow decoder instead of the conventional Transformer architecture with balanced encoder-decoder depth to reduce the computational cost during inference. Experiments in both English and Chinese GEC benchmarks show that aggressive decoding could yield the same predictions as greedy decoding but with a significant speedup for online inference. Its combination with the shallow decoder 
could offer an even higher online inference speedup over the powerful Transformer baseline without quality loss. Not only does our approach allow a single model to achieve the state-of-the-art results in English GEC benchmarks: 66.4 $F_{0.5}$ in the CoNLL-14 and 72.9 $F_{0.5}$ in the BEA-19 test set with an almost $10\times$ online inference speedup over the Transformer-big model, but also it is easily adapted to other languages. Our code is available at \url{https://github.com/AutoTemp/Shallow-Aggressive-Decoding}.

\end{abstract}

\begin{figure*}[t]
    \centering
    \includegraphics[width=\textwidth]{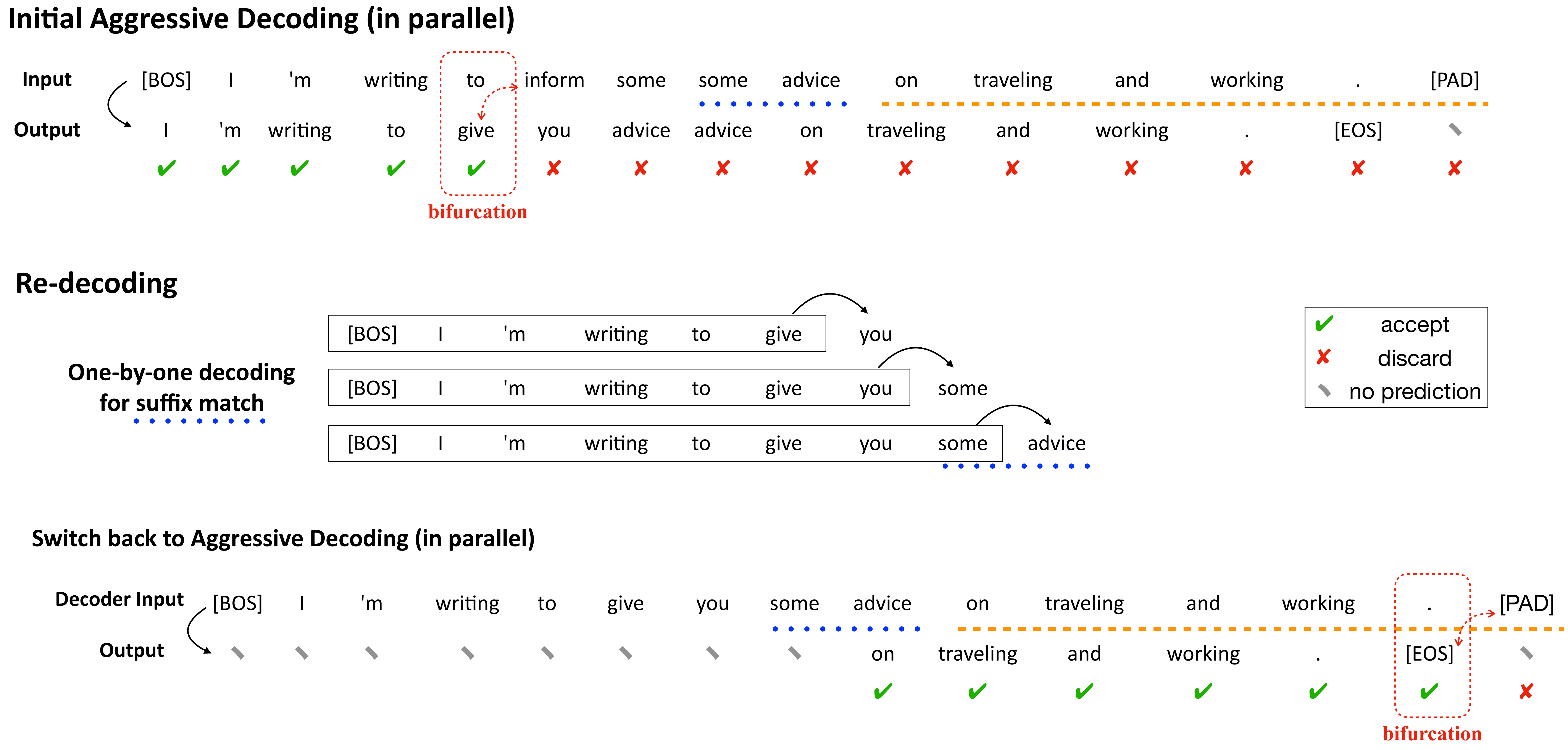} 
    \caption{The overview of aggressive decoding. Aggressive decoding tries decoding as many tokens as possible in parallel with the assumption that the input and output should be almost the same in GEC. When we find a bifurcation between the input and the output of aggressive decoding, then we accept the predictions before (including) the bifurcation, and discard all the predictions after the bifurcation and re-decode them using original one-by-one autoregressive decoding. If we find a suffix match (i.e., \textit{some advice} highlighted with the \textcolor{blue}{blue dot lines}) between the output and the input during one-by-one re-decoding, we switch back to aggressive decoding by copying the tokens (highlighted with the \textcolor{orange}{orange dashed lines}) following the matched tokens in the input to the decoder input by assuming they are likely to be the same.}
    \label{fig:main_flg}
\end{figure*}

\section{Introduction}

The Transformer~\cite{vaswani2017attention} has become the most popular model for Grammatical Error Correction (GEC). In practice, however, the sequence-to-sequence (seq2seq) approach has been blamed recently~\cite{chen2020improving,stahlberg2020seq2edits,omelianchuk2020gector} for its poor inference efficiency in modern writing assistance applications (e.g., Microsoft Office Word\footnote{\url{https://www.microsoft.com/en-us/microsoft-365/word}}, Google Docs\footnote{\url{https://www.google.com/docs/about}} and Grammarly\footnote{\url{https://www.grammarly.com}}) where a GEC model usually performs online inference, instead of batch inference, for proactively and incrementally checking a user's latest completed sentence to offer instantaneous feedback.

To better exploit the Transformer for instantaneous GEC in practice, we propose a novel approach -- Shallow Aggressive Decoding (SAD) to improve the model's online inference efficiency. The core innovation of SAD is aggressive decoding: instead of sequentially decoding only one token at each step, aggressive decoding tries to decode as many tokens as possible in parallel with the assumption that the output sequence should be almost the same with the input. As shown in Figure~\ref{fig:main_flg}, if the output prediction at each step perfectly matches its counterpart in the input sentence, the inference will finish, meaning that the model will keep the input untouched without editing; if the output token at a step does not match its corresponding token in the input, we will discard all the predictions after the bifurcation position and re-decode them in the original autoregressive decoding manner until we find a new opportunity for aggressive decoding. In this way, we can decode the most text in parallel in the same prediction quality as autoregressive greedy decoding, but largely improve the inference efficiency. 

In addition to aggressive decoding, SAD proposes to use a shallow decoder, instead of the conventional Transformer with balanced encoder-decoder depth, to reduce the computational cost for further accelerating inference. 
The experimental results in both English and Chinese GEC benchmarks show that both aggressive decoding and the shallow decoder can significantly improve online inference efficiency. 
By combining these two techniques, our approach shows a $9\times\sim12\times$ online inference speedup over the powerful Transformer baseline without sacrificing the quality.

The contributions of this paper are two-fold:
\begin{itemize}
    \item We propose a novel aggressive decoding approach, allowing us to decode as many token as possible in parallel, which yields the same predictions as greedy decoding but with a substantial improvement of computational parallelism and online inference efficiency.
    \item We propose to combine aggressive decoding with the Transformer with a shallow decoder. Our final approach not only  advances the state-of-the-art in English GEC benchmarks with an almost $10\times$ online inference speedup but also is easily adapted to other languages.
\end{itemize}

\section{Background: Transformer}\label{sec:background}
The Transformer is a seq2seq neural network architecture based on multi-head attention mechanism, which has become the most successful and widely used seq2seq models in various generation tasks such as machine translation, abstractive summarization as well as GEC. 

The original Transformer follows the balanced encoder-decoder architecture: its encoder, consisting of a stack of identical encoder layers, maps an input sentence $\boldsymbol{x}=(x_1, \dots, x_n)$ to a sequence of continuous representation $\boldsymbol{z}=(z_1, \dots, z_n)$; and its decoder, which is composed of a stack of the same number of identical decoder layers as the encoder, generates an output sequence $\boldsymbol{o}=(o_1, \dots, o_m)$ given $\boldsymbol{z}$. 

In the training phase, the model learns an autoregressive scoring model $P(\boldsymbol{y}~|~\boldsymbol{x};\boldsymbol{\Phi})$, implemented with teacher forcing:
\begin{equation}\label{eq:train}
\begin{split}
    \boldsymbol{\Phi^*} & =\arg \max_{\boldsymbol{\Phi}} \log P(\boldsymbol{y}~|~\boldsymbol{x}; \boldsymbol{\Phi}) \\
    & = \arg \max_{\boldsymbol{\Phi}} \sum_{i=0}^{l-1} \log P(y_{i+1}~|~\boldsymbol{y}_{\le i}, \boldsymbol{x}; \boldsymbol{\Phi})
\end{split}
\end{equation}
where $\boldsymbol{y}=(y_1, \dots, y_l)$ is the ground-truth target sequence and $\boldsymbol{y}_{\le i}=(y_0, \dots, y_i)$. As ground truth is available during training, Eq (\ref{eq:train}) can be efficiently obtained as the probability $ P(y_{i+1}~|~\boldsymbol{y}_{\le i}, \boldsymbol{x})$ at each step can be computed in parallel.

During inference, the output sequence $\boldsymbol{o}=(o_1, \dots, o_m)$ is derived by maximizing the following equation:

\begin{equation}
\begin{split}
\boldsymbol{o^*} & = \arg \max_{\boldsymbol{o}} \log P(\boldsymbol{o}~|~\boldsymbol{x}; \boldsymbol{\Phi}) \\
& = \arg \max_{\boldsymbol{o}} \sum_{j=0}^{m-1} \log P(o_{j+1}~|~\boldsymbol{o}_{\le j}, \boldsymbol{x}; \boldsymbol{\Phi})
\end{split}
\end{equation}

Since no ground truth is available in the inference phase, the model has to decode only one token at each step conditioning on the previous decoded tokens $\boldsymbol{o}_{\le j}$ instead of decoding in parallel as in the training phase.

\section{Shallow Aggressive Decoding}

\subsection{Aggressive Decoding}

As introduced in Section \ref{sec:background}, the Transformer decodes only one token at each step during inference. The autoregressive decoding style is the main bottleneck of inference efficiency because it largely reduces computational parallelism.

For GEC, fortunately, the output sequence is usually very similar to the input with only a few edits if any. This special characteristic of the task makes it unnecessary to follow the original autoregressive decoding style; instead, we propose a novel decoding approach -- aggressive decoding which tries to decode as many tokens as possible during inference. The overview of aggressive decoding is shown in Figure \ref{fig:main_flg}, and we will discuss it in detail in the following sections.

\subsubsection{Initial Aggressive Decoding} \label{subsubsec:iag}
The core motivation of aggressive decoding is the assumption that the output sequence $\boldsymbol{o}=(o_1, \dots, o_m)$ should be almost the same with the input sequence $\boldsymbol{x}=(x_1, \dots, x_n)$ in GEC. At the initial step, instead of only decoding the first token $o_1$ conditioning on the special [\textit{BOS}] token $o_0$, aggressive decoding decodes $\boldsymbol{o}_{1 \dots n}$ conditioning on the pseudo previous decoded tokens $\boldsymbol{\hat{o}}_{0 \dots n-1}$ in parallel with the assumption that $\boldsymbol{\hat{o}}_{0 \dots n-1} = \boldsymbol{x}_{0, \dots, n-1}$. Specifically, for $j \in \{0, 1, \dots, n-2, n-1\}$, $o_{j+1}$ is decoded as follows:

\begin{equation}
\begin{split}
o_{j+1}^* &= \arg \max_{o_{j+1}} \log P(o_{j+1}~|\boldsymbol{o}_{\le j}, \boldsymbol{x}; \boldsymbol{\Phi}) \\ & = \arg \max_{o_{j+1}} \log P(o_{j+1}~|~\boldsymbol{\hat{o}}_{\le j}, \boldsymbol{x}; \boldsymbol{\Phi}) \\ 
& = \arg \max_{o_{j+1}} \log P(o_{j+1}~|~\boldsymbol{x}_{\le j}, \boldsymbol{x}; \boldsymbol{\Phi})
\end{split}
\end{equation}
where $\boldsymbol{\hat{o}}_{\le j}$ is the pseudo previous decoded tokens at step $j+1$, which is assumed to be the same with $\boldsymbol{x}_{\le j}$.

After we obtain $\boldsymbol{o}_{1...n}$, we verify whether $\boldsymbol{o}_{1...n}$ is actually identical to $\boldsymbol{x}_{1...n}$ or not. If $\boldsymbol{o}_{1...n}$ is fortunately exactly the same with $\boldsymbol{x}_{1...n}$, the inference will finish, meaning that the model finds no grammatical errors in the input sequence $\boldsymbol{x}_{1...n}$ and keeps the input untouched. In more cases, however, $\boldsymbol{o}_{1...n}$ will not be exactly the same with $\boldsymbol{x}_{1...n}$. In such a case, we have to stop aggressive decoding and find the first bifurcation position $k$ so that $\boldsymbol{o}_{1...k-1} = \boldsymbol{x}_{1...k-1}$ and $o_k \neq x_k$.

Since $\boldsymbol{o}_{1...k-1} = \boldsymbol{\hat{o}}_{1...k-1} = \boldsymbol{x}_{1...k-1}$, the predictions $\boldsymbol{o}_{1...k}$ could be accepted as they will not be different even if they are decoded through the original autoregressive greedy decoding. However, for the predictions $\boldsymbol{o}_{k+1 \dots n}$, we have to discard and re-decode them because $o_k \neq \hat{o}_k$.

\begin{algorithm*}[t]
\caption{Aggressive Decoding\label{alg:ag}}
\textbf{Input:} $\boldsymbol{\Phi}$, $\boldsymbol{x}=(\textrm{[}BOS\textrm{]}, x_1, \dots, x_n, \textrm{[}PAD\textrm{]})$, $\boldsymbol{o}=(o_0)=(\textrm{[}BOS\textrm{]})$; \\
\textbf{Output:} $\boldsymbol{o}_{1 \dots j}=(o_1, \dots, o_j)$;
\begin{algorithmic}[1]
\State Initialize $j \gets 0$;
\While{$o_j \neq$ [$EOS$] and $j <$ MAX\_LEN}
\If{$\boldsymbol{o}_{j-q \dots j}$ $(q \ge 0)$ is a unique substring of $\boldsymbol{x}$ such that $\exists~!~i:   \boldsymbol{o}_{j-q \dots j} = \boldsymbol{x}_{i-q \dots i}$}
\State{Aggressive Decode $\boldsymbol{\widetilde{o}}_{j+1 \dots}$ according to Eq (\ref{eq:agcore}) and Eq (\ref{eq:agcore1})};
\State Find bifurcation $j+k$ ($k>0$) such that $\boldsymbol{\widetilde{o}}_{j+1 \dots j+k-1} = \boldsymbol{x}_{i+1 \dots i+k-1}$ and ${\widetilde{o}}_{j+k} \neq x_{i+k}$;
\State $\boldsymbol{o} \gets \textsc{Cat}(\boldsymbol{o}, \boldsymbol{\widetilde{o}}_{j+1 \dots j+k})$;
\State $j \gets j+k$;
\Else 
\State Decode $o_{j+1}^* = \arg \max_{o_{j+1}} P(o_{j+1}~|~\boldsymbol{o}_{\le j}, \boldsymbol{x}; \boldsymbol{\Phi})$;
\State $\boldsymbol{o} \gets \textsc{Cat}(\boldsymbol{o}, o_{j+1}^*)$;
\State $j \gets j+1$;
\EndIf
\EndWhile
\end{algorithmic}
\end{algorithm*}

\subsubsection{Re-decoding} \label{subsubsec:rdra}  
As $o_k \neq \hat{o}_k = x_k$, we have to re-decode for $o_{j+1}$ ($j \ge k$) one by one following the original autoregressive decoding:

\begin{equation}
o_{j+1}^*= \arg \max_{o_{j+1}} P(o_{j+1}~|~\boldsymbol{o}_{\le j}, \boldsymbol{x}; \boldsymbol{\Phi})
\end{equation}

After we obtain $\boldsymbol{o}_{\le j}$ ($j>k$), we try to match its suffix to the input sequence $\boldsymbol{x}$ for further aggressive decoding. If we find its suffix $\boldsymbol{o}_{j-q \dots j}$ ($q \ge 0$) is the unique substring of $\boldsymbol{x}$ such that $\boldsymbol{o}_{j-q \dots j}=\boldsymbol{x}_{i-q \dots i}$, then we can assume that $\boldsymbol{o}_{j+1 \dots}$ will be very likely to be the same with $\boldsymbol{x}_{i+1 \dots}$ because of the special characteristic of the task of GEC.

If we fortunately find such a suffix match, then we can switch back to aggressive decoding to decode in parallel with the assumption $\boldsymbol{\hat{o}}_{j+1 \dots} = \boldsymbol{x}_{i+1 \dots}$. Specifically, the token $o_{j+t}$ ($t>0$) is decoded as follows:

\begin{equation} \label{eq:agcore}
    o_{j+t}^* = \arg \max_{o_{j+t}} P(o_{j+t}~|~\boldsymbol{o}_{< j+t}, \boldsymbol{x}; \boldsymbol{\Phi}) 
\end{equation}
In Eq (\ref{eq:agcore}), $\boldsymbol{o}_{<j+t}$ is derived as follows:
\begin{equation} \label{eq:agcore1}
\begin{split}
\boldsymbol{o}_{<j+t} & = \textsc{Cat}(\boldsymbol{o}_{\le j}, \boldsymbol{\hat{o}}_{j+1 \dots j+t-1}) \\
 & = \textsc{Cat}(\boldsymbol{o}_{\le j}, \boldsymbol{x}_{i+1 \dots i+t-1}) 
\end{split}
\end{equation}
where \textsc{Cat}($\boldsymbol{a}$, $\boldsymbol{b}$) is the operation that concatenates two sequences $\boldsymbol{a}$ and $\boldsymbol{b}$.

Otherwise (i.e., we cannot find a suffix match at the step), we continue decoding using the original autoregressive greedy decoding approach until we find a suffix match.

We summarize the process of aggressive decoding in Algorithm \ref{alg:ag}. For simplifying implementation, we make minor changes in Algorithm \ref{alg:ag}: 1) we set $o_0=x_0=\textrm{[}BOS\textrm{]}$ in Algorithm \ref{alg:ag}, which enables us to regard the initial aggressive decoding as the result of suffix match of $o_0=x_0$; 2) we append a special token $\textrm{[}PAD\textrm{]}$ to the end of $\boldsymbol{x}$ so that the bifurcation (in the $5^{th}$ line in Algorithm \ref{alg:ag}) must exist (see the bottom example in Figure \ref{fig:main_flg}). Since we discard all the computations and predictions after the bifurcation for re-decoding, aggressive decoding guarantees that generation results are exactly the same as greedy decoding (i.e., beam=1). However, as aggressive decoding decodes many tokens in parallel, it largely improves the computational parallelism during inference, greatly benefiting the inference efficiency.

\subsection{Shallow Decoder}\label{subsec:sd}
Even though aggressive decoding can significantly improve the computational parallelism during inference, it inevitably leads to intensive computation and even possibly introduces additional computation caused by re-decoding for the discarded predictions. 

To reduce the computational cost for decoding, we propose to use a shallow decoder, which has proven to be an effective strategy~\cite{kasai2020deep, li2021efficient} in neural machine translation (NMT), instead of using the Transformer with balanced encoder-decoder depth as the previous state-of-the-art Transformer models in GEC. By combining aggressive decoding with the shallow decoder, we are able to further improve the inference efficiency. 

\section{Experiments}

\begin{table*}[t] 
\centering
\scalebox{0.95}{
\begin{tabular}{l|c|c|c|ccc}
\hline
\multirow{2}{*}{\bf Model}       & \multirow{2}{*}{\bf Synthetic Data} & \multirow{2}{*}{\bf Total Latency (s)} & \multirow{2}{*}{\bf Speedup} & \multicolumn{3}{c}{ \textbf{CoNLL-13}} \\  
 & & &  & $P$ & $R$ & $F_{0.5}$ \\ \hline
Transformer-big (beam=5) & No & 440 & 1.0$\times$ & \bf 53.84 & 18.00 & \bf 38.50 \\
Transformer-big (greedy) & No & 328 & 1.3$\times$ & 52.75 & \bf 18.34 & 38.36 \\
Transformer-big (aggressive) & No & \bf 54 & \bf \boldmath 8.1$\times$ & 52.75 & \bf 18.34 & 38.36 \\ \hline
Transformer-big (beam=5) & Yes & 437 & 1.0$\times$ & \bf 57.06 & 23.62 &  44.47 \\
Transformer-big (greedy) & Yes & 320 & 1.4$\times$ & 56.45 & \bf 24.70 & \bf 44.91     \\
Transformer-big (aggressive) & Yes & \bf 60 & \bf \boldmath 7.3$\times$ & 56.45 & \bf 24.70 & \bf 44.91     \\ \hline
\end{tabular}
}
\caption{The performance and online inference efficiency of the Transformer-big with aggressive decoding in our validation set (CoNLL-13) that contains 1,381 sentences. We use Transformer-big (beam=5) as the baseline to compare the performance and efficiency of aggressive decoding. \label{tab:agresult1}}
\end{table*}

\subsection{Data and Model Configuration}
We follow recent work in English GEC to conduct experiments in the restricted training setting of BEA-2019 GEC shared task~\cite{bryant2019bea}: We use Lang-8 Corpus of Learner English~\cite{mizumoto2011mining}, NUCLE~\cite{dahlmeier2013building}, FCE~\cite{yannakoudakis2011new} and W\&I+LOCNESS~\cite{granger1998computer, bryant2019bea} as our GEC training data. 
For facilitating fair comparison in the efficiency evaluation, we follow the previous studies~\cite{omelianchuk2020gector,chen2020improving} which conduct GEC efficiency evaluation to use CoNLL-2014~\cite{ng2014conll} dataset that contains 1,312 sentences as our main test set, and evaluate the speedup as well as Max-Match~\cite{dahlmeier2012better} precision, recall and $F_{0.5}$ using their official evaluation scripts\footnote{\url{https://github.com/nusnlp/m2scorer}}. For validation, we use CoNLL-2013~\cite{ng-etal-2013-conll} that contains 1,381 sentences as our validation set.
We also test our approach on NLPCC-18 Chinese GEC shared task~\cite{zhao2018overview}, following their training\footnote{Following~\newcite{chen2020improving}, we sample 5,000 training instances as the validation set.} and evaluation setting, to verify the effectiveness of our approach in other languages. To compare with the state-of-the-art approaches in English GEC that pretrain with synthetic data, we also synthesize 300M error-corrected sentence pairs for pretraining the English GEC model following the approaches of \newcite{grundkiewicz2019neural} and \newcite{zhang2019sequence}. Note that in the following evaluation sections, the models evaluated are by default trained without the synthetic data unless they are explicitly mentioned.   

We use the most popular GEC model architecture -- Transformer (big) model~\cite{vaswani2017attention} as our baseline model which has a 6-layer encoder and 6-layer decoder with 1,024 hidden units. We train the English GEC model using an encoder-decoder shared vocabulary of 32K Byte Pair Encoding~\cite{sennrich2016neural} tokens and train the Chinese GEC model with 8.4K Chinese characters. We include more training details in the supplementary notes. For inference, we use greedy decoding\footnote{Our implementation of greedy decoding is simplified for higher efficiency ($1.3\times\sim1.4\times$ speedup over beam=5) than the implementation of beam=1 decoding in fairseq (around $1.1\times$ speedup over beam=5).} by default.

All the efficiency evaluations are conducted in the online inference setting (i.e., batch size=1) as we focus on instantaneous GEC. We perform model inference with fairseq\footnote{\url{https://github.com/pytorch/fairseq}} implementation using Pytorch 1.5.1 with 1 Nvidia Tesla V100-PCIe of 16GB GPU memory under CUDA 10.2.

\begin{table*}[t] 
\centering
\small
\begin{tabular}{c|c|p{5.5cm}|p{6cm}}
\hline
\bf Speedup & \bf Edit Ratio & \bf Input & \bf Output \\
 \hline
16.7$\times$ & 0 & Personally , I think surveillance technology such as RFID ( radio-frequency identification ) should not be used to track people , for the benefit it brings to me can not match the concerns it causes . & [\textcolor{blue}{Personally , I think surveillance technology such as RFID ( radio-frequency identification ) should not be used to track people , for the benefit it brings to me can not match the concerns it causes .}]$_0$ \\
 \hline
 5.8$\times$ & 0 & Nowadays , people use the all-purpose smart phone for communicating . & [\textcolor{blue}{Nowadays , people use the all-purpose smart phone for communicating .}]$_0$ \\ 
 \hline
 6.8$\times$ & 0.03 & Because that the birth rate is reduced while the death rate is also reduced , the percentage of the elderly is increased while that of the youth is decreased . & [\textcolor{blue}{Because the}]$_0$ [\textcolor{red}{birth}]$_1$ [\textcolor{blue}{rate is reduced while the death rate is also reduced , the percentage of the elderly is increased while that of the youth is decreased .}]$_2$ \\
 \hline
 5.1$\times$ & 0.06 & More importantly , they can share their ideas of how to keep healthy through Internet , to make more interested people get involve and find ways to make life longer and more wonderful . & [\textcolor{blue}{More importantly , they can share their ideas of how to keep healthy through the}]$_0$ [\textcolor{red}{Internet}]$_1$ [\textcolor{blue}{, to make more interested people get involved}]$_2$ [\textcolor{red}{and}]$_3$ [\textcolor{red}{find}]$_4$ [\textcolor{blue}{ways to make life longer and more wonderful .}]$_5$   \\
 \hline
 3.5$\times$ & 0.13 & As a result , people have more time to enjoy advantage of modern life . &  [\textcolor{blue}{As a result , people have more time to enjoy the}]$_0$ [\textcolor{red}{advantages}]$_1$ [\textcolor{red}{of}]$_2$ [\textcolor{blue}{modern life .}]$_3$  \\
 \hline
 1.5$\times$ & 0.27 & Nowadays , technology is more advance than the past time . & [\textcolor{blue}{Nowadays , technology is more advanced}]$_0$ [\textcolor{red}{than}]$_1$ [\textcolor{blue}{in}]$_2$ [\textcolor{red}{the}]$_3$ [\textcolor{blue}{past .}]$_4$ \\
 \hline
 1.4$\times$ & 0.41 & People are able to predicate some disasters like the earth quake and do the prevention beforehand . & [\textcolor{blue}{People are able to predict}]$_0$ [\textcolor{red}{disasters}]$_1$ [\textcolor{blue}{like the earthquake}]$_2$ [\textcolor{red}{and}]$_3$ [\textcolor{blue}{prevent}]$_4$ [\textcolor{red}{them}]$_5$ [\textcolor{red}{beforehand}]$_6$ [\textcolor{blue}{.}]$_7$\\
  \hline
\end{tabular}
\caption{Examples of various speedup ratios by aggressive decoding over greedy decoding in CoNLL-13. We show how the examples are decoded in the column of \textbf{Output}, where the tokens within a \textcolor{blue}{blue block} are decoded in parallel through aggressive decoding while the tokens in \textcolor{red}{red blocks} are decoded through the original autoregressive greedy decoding.\label{tab:casestudy}}
\end{table*}

\subsection{Evaluation for Aggressive Decoding}

We evaluate aggressive decoding in our validation set (CoNLL-13) which contains 1,381 validation examples. As shown in Table \ref{tab:agresult1}, aggressive decoding achieves a $7\times\sim8\times$ speedup over the original autoregressive beam search (beam=5), and generates exactly the same predictions as greedy decoding, as discussed in Section \ref{subsubsec:rdra}. Since greedy decoding can achieve comparable overall performance (i.e., $F_{0.5}$) with beam search while it tends to make more edits resulting in higher recall but lower precision, the advantage of aggressive decoding in practical GEC applications is obvious given its strong performance and superior efficiency.

\begin{figure}[t]
    \centering
    \includegraphics[width=8.3cm]{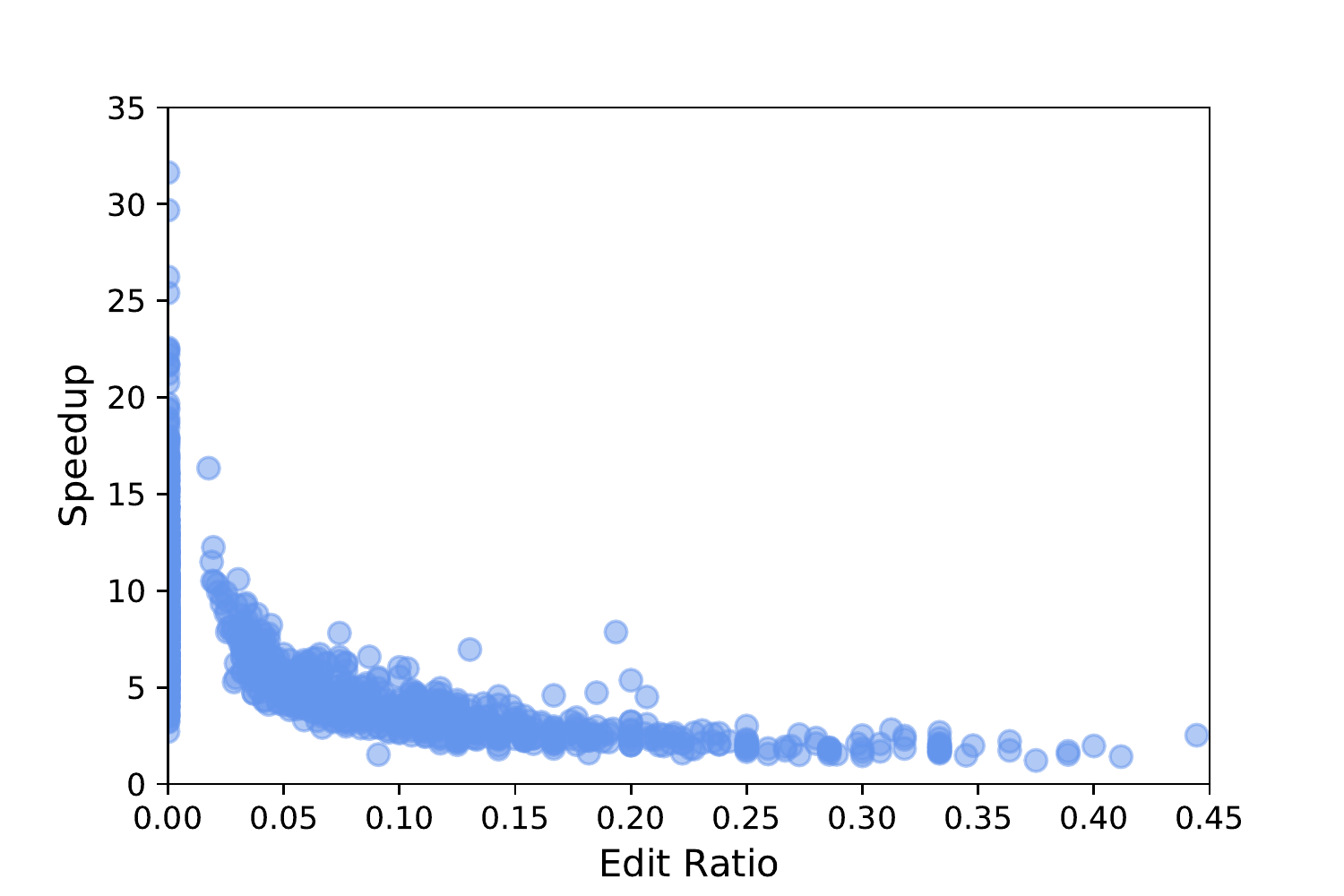}
    \caption{The speedup (over greedy decoding) distribution of all the 1,381 validation examples with respect to their edit ratio in CoNLL-13.}
    \label{fig:density}
\end{figure}

We further look into the efficiency improvement by aggressive decoding. Figure \ref{fig:density} shows the speedup distribution of the 1,381 examples in CoNLL-13 with respect to their edit ratio which is defined as the normalized (by the input length) edit distance between the input and output. It is obvious that the sentences with fewer edits tend to achieve higher speedup, which is consistent with our intuition that most tokens in such sentences can be decoded in parallel through aggressive decoding; on the other hand, for the sentences that are heavily edited, their speedup is limited because of frequent re-decoding. To give a more intuitive analysis, we also present concrete examples with various speedup in our validation set to understand how aggressive decoding improves the inference efficiency in Table \ref{tab:casestudy}.

\begin{table}[t]
\begin{tabular}{c|c|c} \hline
\textbf{$L_{max}$} & \textbf{Total Latency (s)} & \textbf{Speedup} \\ \hline
1 (Baseline) & 328 & 1.0$\times$\\ 
2 & 208 & 1.6$\times$ \\
3 & 148 & 2.2$\times$ \\
5 & 109 & 3.0$\times$ \\
10 & 75 & 4.4$\times$ \\
20 & 64 & 5.1$\times$ \\ 
40 & \bf 54 & \bf \boldmath 6.1$\times$ \\ 
Unlimited & \bf 54 & \bf \boldmath 6.1$\times$ \\ \hline
\end{tabular}
\caption{The ablation study of the effect of constraining the maximal aggressive decoding length $L_{max}$ on the online inference efficiency in CoNLL-13. Note that in CoNLL-13, the average length of an example is 21 and 96\% examples are shorter than 40 tokens.}\label{tab:maxlen}
\end{table}

\begin{table}[t]
\centering
\scalebox{0.85}
{
\begin{tabular}{c|c|c|c} \hline
\multirow{2}{*}{\textbf{\begin{tabular}[c]{@{}c@{}}Model\\ (Enc+Dec)\end{tabular}}} & \multicolumn{1}{c|}{\textbf{CoNLL-13}} & \multirow{2}{*}{\textbf{\begin{tabular}[c]{@{}c@{}}Total\\ Latency\end{tabular}}} & \multirow{2}{*}{\textbf{Speedup}} \\ 
 & $F_{0.5}$ & & \\ \hline
6+6 & 38.36 & 328 & 1.0$\times$ \\ \hline
3+6 & 36.26 & 314 & 1.0$\times$ \\
9+6 & 38.82 & 345 & 1.0$\times$ \\ \hline
6+3 & 37.95 & 175 & 1.9$\times$ \\
6+9 & 38.02 & 457 & 0.7$\times$ \\ \hline 
7+5 & 38.49 & 271 & 1.2$\times$ \\
8+4 & 38.63 & 240 & 1.4$\times$ \\ 
9+3 & \bf 38.88 & 181 & 1.8$\times$ \\ 
10+2 & 38.21 & 137 & 2.4$\times$ \\ 
11+1 & 38.15 & \bf 86 & \bf \boldmath 3.8$\times$ \\ \hline 
\end{tabular}
}
\caption{The performance and efficiency of the Transformer with different encoder and decoder depths in CoNLL-13, where 6+6 is the original Transformer-big model that has a 6-layer encoder and a 6-layer decoder.}\label{tab:depth}
\end{table}

Moreover, we conduct an ablation study to investigate whether it is necessary to constrain the maximal aggressive decoding length\footnote{Constraining the maximal aggressive decoding length to $L_{max}$ means that the model can only aggressively decode at most $L_{max}$ tokens in parallel.}, because it might become highly risky to waste large amounts of computation because of potential re-decoding for a number of steps after the bifurcation if we aggressively decode a very long sequence in parallel. Table \ref{tab:maxlen} shows the online inference efficiency with different maximal aggressive decoding lengths. It appears that constraining the maximal aggressive decoding length does not help improve the efficiency; instead, it slows down the inference if the maximal aggressive decoding length is set to a small number. We think the reason is that sentences in GEC datasets are rarely too long. For example, the average length of the sentences in CoNLL-13 is 21 and 96\% of them are shorter than 40 tokens. Therefore, it is unnecessary to constrain the maximal aggressive decoding length in GEC.

\begin{table*}[t]
\centering
\small
\begin{tabular}{l|c|c|cccc}
\hline
\multirow{2}{*}{\textbf{Model}} & \multirow{2}{*}{\textbf{Synthetic Data}} & \bf Multi-stage & \multicolumn{4}{c}{\textbf{CoNLL-14}} \\ 
 & & \bf Fine-tuning & $P$ & $R$ & $F_{0.5}$ & Speedup \\ 
 \hline
\it Transformer-big (beam=5) & No & No & 60.2 & 32.1 & \bf 51.2 & 1.0$\times$ \\
\textit{Levenshtein Transformer}$^\star$~\cite{gu2019levenshtein} & No & No & 53.1 & 23.6 & 42.5 & 2.9$\times$ \\
\textit{LaserTagger}$^\star$~\cite{malmi2019encode} & No & No & 50.9 & 26.9 & 43.2 & \bf \boldmath \underline{29.6$\times$} \\
\textit{Span Correction}$^\star$~\cite{chen2020improving} & No & No & \bf 66.0 & 24.7 & 49.5 & 2.6$\times$ \\
Our approach (9+3) & No & No & 58.8 & \bf 33.1 & 50.9 & \bf \boldmath 10.5$\times$ \\ 
 \hline
\it Transformer-big (beam=5) & Yes & No & 73.0 &  38.1 & 61.6 & 1.0$\times$ \\
\textit{PIE}$^\star$~\cite{awasthi2019parallel} & Yes & No & 66.1 & \bf 43.0 & 59.7 & \bf \boldmath \underline{10.3$\times$} \\
\textit{Span Correction}$^\star$~\cite{chen2020improving}  & Yes & No & 72.6 & 37.2 & 61.0 & 2.6$\times$ \\
Our approach (9+3) & Yes & No & \bf 73.3 & 41.3 & \bf 63.5 & \bf \boldmath 10.3$\times$ \\ \hline
\textit{Seq2Edits}~\cite{stahlberg2020seq2edits} & Yes & Yes & 63.0 & 45.6 & 58.6 & - \\
\textit{GECToR}(\textit{RoBERTa})~\cite{omelianchuk2020gector} & Yes & Yes &  73.9 &  41.5 &  64.0 & \bf \boldmath 12.4$\times$ \\
\textit{GECToR}(\textit{XLNet})~\cite{omelianchuk2020gector} & Yes & Yes & \bf 77.5 &  40.1 & 65.3 &  - \\
Our approach (12+2 BART-Init) & Yes & Yes & 71.0 & \bf 52.8 & \bf 66.4 &  9.6$\times$ \\ 
\hline
\end{tabular}
\caption{The performance and online inference efficiency evaluation of efficient GEC models in CoNLL-14. For the models with $\star$, their performance and speedup numbers are from \newcite{chen2020improving} who evaluate the online efficiency in the same runtime setting (e.g., GPU and runtime libraries) with ours. The underlines indicate the speedup numbers of the models are evaluated with Tensorflow based on their released codes, which are not strictly comparable here. Note that for \textit{GECToR}, we re-implement its inference process of \textit{GECToR} (RoBERTa) using fairseq for testing its speedup in our setting. - means the speedup cannot be tested in our runtime environment because the model has not been released or not implemented in fairseq.} \label{tab:finalenglish}
\end{table*}

\subsection{Evaluation for Shallow Decoder}

We study the effects of changing the number of encoder and decoder layers in the Transformer-big on both the performance and the online inference efficiency. By comparing 6+6 with 3+6 and 9+6 in Table \ref{tab:depth}, we observe the performance improves as the encoder becomes deeper, demonstrating the importance of the encoder in GEC. In contrast, by comparing the 6+6 with 6+3 and 6+9, we do not see a substantial fluctuation in the performance, indicating no necessity of a deep decoder. Moreover, it is observed that a deeper encoder does not significantly slow down the inference but a shallow decoder can greatly improve the inference efficiency. This is because Transformer encoders can be parallelized efficiently on GPUs, whereas Transformer decoders are auto-regressive and hence the number of layers greatly affects decoding speed, as discussed in Section \ref{subsec:sd}. These observations motivate us to make the encoder deeper and the decoder shallower. 

As shown in the bottom group of Table \ref{tab:depth}, we try different combinations of the number of encoder and decoder layers given approximately the same parameterization budget as the Transformer-big. It is interesting to observe that 7+5, 8+4 and 9+3 achieve the comparable and even better performance than the Transformer-big baseline with much less computational cost. When we further increase the encoder layer and decrease the decoder layer, we see a drop in the performance of 10+2 and 11+1 despite the improved efficiency because it becomes difficult to train the Transformer with extremely imbalanced encoder and decoder well, as indicated\footnote{They show that sequence-level knowledge distillation (KD) may benefit training the extremely imbalanced Transformer in NMT. However, we do not conduct KD for fair comparison to other GEC models in previous work.} by the previous work~\cite{kasai2020deep,li2021efficient,gu2020fully}.


Since the 9+3 model achieves the best result with an around $2\times$ speedup in the validation set with almost the same parameterization budget, we choose it as the model architecture to combine with aggressive decoding for final evaluation.

\begin{table*}[t]
\centering
\begin{tabular}{l|cccc}
\hline
\multirow{2}{*}{\textbf{Model}} &  \multicolumn{4}{c}{\textbf{NLPCC-18}} \\ 
 & $P$ & $R$ & $F_{0.5}$ & Speedup \\ \hline
\it Transformer-big (beam=5) & 36.0 &  17.2 & \bf 29.6 & 1.0$\times$ \\
\textit{Levenshtein Transformer}$^\star$ & 24.9 & 15.0 & 22.0 & 3.1$\times$ \\
\textit{LaserTagger}$^\star$ & 25.6 & 10.5 & 19.9 & \bf \boldmath \underline{38.0$\times$} \\
\textit{Span Correction}$^\star$ & \bf 37.3 & 14.5 & 28.4 & 2.7$\times$ \\ \hline
Our approach (9+3) & 33.0 & \bf 20.5 & 29.4 & \bf \boldmath 12.0$\times$ \\ \hline
\end{tabular}
\caption{The performance and online inference efficiency evaluation for the language-independent efficient GEC models in the NLPCC-18 Chinese GEC benchmark.}\label{tab:chinese}
\end{table*}

\subsection{Results}
We evaluate our final approach -- shallow aggressive decoding which combines aggressive decoding with the shallow decoder. Table~\ref{tab:finalenglish} shows the performance and efficiency of our approach and recently proposed efficient GEC models that are all faster than the Transformer-big baseline in CoNLL-14 test set. Our approach (the 9+3 model with aggressive decoding) that is pretrained with synthetic data achieves 63.5 $F_{0.5}$ with $10.3\times$ speedup over the Transformer-big baseline, which outperforms the majority\footnote{It is notable that \textit{PIE} is not strictly comparable here because their training data is different from ours: \textit{PIE} does not use the W\&I+LOCNESS corpus.} of the efficient GEC models in terms of either quality or speed. The only model that shows advantages over our 9+3 model is \textit{GECToR} which is developed based on the powerful pretrained models (e.g., RoBERTa~\cite{liu2019roberta} and XLNet~\cite{yang2019xlnet}) with its multi-stage training strategy. Following \textit{GECToR}'s recipe, we leverage the pretrained model BART~\cite{lewis2019bart} to initialize a 12+2 model which proves to work well in NMT~\cite{li2021efficient} despite more parameters, and apply the multi-stage fine-tuning strategy used in \newcite{stahlberg2020seq2edits}. The final single model\footnote{The same model checkpoint also achieves the state-of-the-art result -- 72.9 $F_{0.5}$ with a $9.3\times$ speedup in the BEA-19 test set.} with aggressive decoding achieves the state-of-the-art result -- 66.4 $F_{0.5}$ in the CoNLL-14 test set with a $9.6\times$ speedup over the Transformer-big baseline. 

Unlike \textit{GECToR} and \textit{PIE} that are difficult to adapt to other languages despite their competitive speed because they are specially designed for English GEC with many manually designed language-specific operations like the transformation of verb forms (e.g.,
VBD$\to$VBZ) and prepositions (e.g., in$\to$at), our approach is data-driven without depending on language-specific features, and thus can be easily adapted to other languages (e.g., Chinese). As shown in Table \ref{tab:chinese}, our approach consistently performs well in Chinese GEC, showing an around $12.0\times$ online inference speedup over the Transformer-big baseline with comparable performance.

\section{Related Work}

The state-of-the-art of GEC has been significantly advanced owing to the tremendous success of seq2seq learning~\cite{sutskever2014sequence} and the Transformer~\cite{vaswani2017attention}. Most recent work on GEC focuses on improving the performance of the Transformer-based GEC models. However, except for the approaches that add synthetic erroneous data for pretraining~\cite{ge-etal-2018-fluency, grundkiewicz2019neural, zhang2019sequence,lichtarge2019corpora,zhou2020improving,wan2020improving}, most methods that improve performance~\cite{ge2018reaching,kaneko2020encoder} introduce additional computational cost and thus slow down inference despite the performance improvement.

To make the Transformer-based GEC model more efficient during inference for practical application scenarios, some recent studies have started exploring the approaches based on edit operations. Among them, PIE \cite{awasthi2019parallel} and GECToR \cite{omelianchuk2020gector} propose to accelerate the inference by simplifying GEC from sequence generation
to iterative edit operation tagging. However, as they rely on many language-dependent edit operations such as the conversion of singular nouns to plurals, it is difficult for them to adapt to other languages. LaserTagger \cite{malmi2019encode} uses the similar method but it is data-driven and language-independent by learning operations from training data. However, its performance is not so desirable as its seq2seq counterpart despite its high efficiency. The only two previous efficient approaches that are both language-independent and good-performing are \newcite{stahlberg2020seq2edits} which uses span-based edit operations to correct sentences to save the time for copying unchanged tokens, and \newcite{chen2020improving} which first identifies incorrect spans with a tagging model then only corrects these spans with a generator. However, all the approaches have to extract edit operations and even conduct token alignment in advance from the error-corrected sentence pairs for training the model. In contrast, our proposed shallow aggressive decoding tries to accelerate the model inference through parallel autoregressive decoding which is related to some previous work~\cite{ghazvininejad2019mask,stern2018blockwise} in neural machine translation (NMT), and the imbalanced encoder-decoder architecture which is recently explored by \newcite{kasai2020deep} and \newcite{li2021efficient} for NMT. Not only is our approach language-independent, efficient and guarantees that its predictions are exactly the same with greedy decoding, but also does not need to change the way of training, making it much easier to train without so complicated data preparation as in the edit operation based approaches.

\section{Conclusion and Future Work}

In this paper, we propose Shallow Aggressive Decoding (SAD) to accelerate online inference efficiency of the Transformer for instantaneous GEC. Aggressive decoding can yield the same prediction quality as autoregressive greedy decoding but with much less latency. Its combination with the Transformer with a shallow decoder can achieve state-of-the-art performance with a $9\times\sim12\times$ online inference speedup over the Transformer-big baseline for GEC.

Based on the preliminary study of SAD in GEC, we plan to further explore the technique for accelerating the Transformer for other sentence rewriting tasks, where the input is similar to the output, such as style transfer and text simplification. We believe SAD is promising to become a general acceleration methodology for writing intelligence models in modern writing assistant applications that require fast online inference.

\section*{Acknowledgments}

We thank all the reviewers for their valuable comments to improve our paper. We thank Xingxing Zhang, Xun Wang and Si-Qing Chen for their insightful discussions and suggestions. The work is supported by National Natural Science Foundation of China under Grant No.62036001. The corresponding author of this paper is Houfeng Wang.


\bibliographystyle{acl_natbib}
\bibliography{acl2021}

\appendix
\section{Hyper-parameters}

Hyper-parameters of training the Transformer for English GEC are listed in table \ref{tab:param}. The hyper-parameters for Chinese GEC are the same with those of training from scratch. 

\begin{table}[h]
\centering
\small
\begin{tabular} {lr} 
\hline
Configurations	         &	Values
\\
\hline 
\multicolumn{2}{c}{\bf Train From Scratch} \\
\hline
Model Architecture 	    & Transformer (big)    \\
                        & ~\cite{vaswani2017attention} \\
Number of epochs		& 60				\\
Devices                 & 4 Nvidia V100 GPU      \\
Max tokens per GPU		& 5120					\\
Update Frequency        & 4                     \\
Optimizer 				& Adam 					\\
						& ($\beta_1$=0.9, $\beta_2$=0.98, $\epsilon$=$1\times10^{-8}$)	\\
                        & ~\cite{kingma2014adam} \\
Learning rate 			& [$3\times10^{-4}$ , $5\times10^{-4}$] \\
Learning rate scheduler & inverse sqrt \\ 
Warmup                  & 4000 \\
Weight decay            & 0.0 \\
Loss Function 			& label smoothed cross entropy \\
						& (label-smoothing=0.1) \\
						& ~\cite{szegedy2016rethinking} \\
Dropout 				& [0.3, 0.4, 0.5] \\
\hline
\multicolumn{2}{c}{\bf Pretrain} \\
\hline
Number of epochs 		& 10					\\
Devices                 & 8 Nvidia V100 GPU      \\
Update Frequency        & 8                     \\
Learning rate 			& $3\times10^{-4}$			\\
Warmup                  & 8000 \\
Dropout					& 0.3	\\
\hline
\multicolumn{2}{c}{\bf Fine-tune} \\
\hline
Number of epochs 		& 60					\\
Devices                 & 4 Nvidia V100 GPU      \\
Update Frequency        & 4                     \\
Learning rate 			& $3\times10^{-4}$			\\
Warmup                  & 4000 \\
Dropout					& 0.3	\\
\hline
\end{tabular}
\caption{Hyper-parameters values of training from scratch, pretraining and fine-tuning. \label{tab:param}} 
\end{table}

\section{CPU Efficiency}


\begin{table}[h]
\centering
\scalebox{0.8}{
\begin{tabular}{c|c|c|c|c}
\hline
\multirow{2}{*}{\textbf{\begin{tabular}[c]{@{}c@{}}Model\\ (Enc+Dec)\end{tabular}}} &
\multirow{2}{*}{\textbf{Thread}} &
\textbf{Beam=5} & \textbf{Greedy} & \textbf{Aggressive} \\
 & & Speedup & Speedup & Speedup \\
 \hline
6+6 & 8 & 1$\times$ & 1.6$\times$ & 6.5$\times$ \\
9+3 & 8 & 1.5$\times$ & 2.5$\times$ & 8.0$\times$ \\
\hline
6+6 & 2 & 1$\times$ & 2.1$\times$ & 6.1$\times$ \\
9+3 & 2 & 1.5$\times$ & 3.1$\times$ & 7.6$\times$ \\
\hline
\end{tabular}
}
\caption{The efficiency of the Transformer with different encoder and decoder depths in CoNLL-13 on CPU with 8 and 2 threads. \label{tab:cpu} }
\end{table}

Table~\ref{tab:cpu} shows total latency and speedup of the Transformer with different encoder-decoder depth on an Intel\textsuperscript{\tiny\textregistered} Xeon\textsuperscript{\tiny\textregistered} E5-2690 v4 Processor(2.60GHz) with 8 and 2 threads\footnote{We set OMP\_NUM\_THREADS to 8 or 2.}, respectively. Our approach achieves a $7\times\sim8\times$ online inference speedup over the Transformer-big baseline on CPU. 
\end{document}